\documentclass[lettersize,journal]{IEEEtran}
\usepackage{amsmath,amsfonts}
\usepackage{algorithmic}
\usepackage{algorithm}
\usepackage{array}
\usepackage[caption=false,font=normalsize,labelfont=sf,textfont=sf]{subfig}
\usepackage{textcomp}
\usepackage{stfloats}
\usepackage{url}
\usepackage{verbatim}
\usepackage{graphicx}
\usepackage{cite}
\usepackage{orcidlink}
\usepackage{multirow}
\usepackage{hyperref}

\begin{document}

\title{Stereo 4D Radar for 3D Object Detection: \\ Integrating Geometric Alignment and Absolute Velocity Estimation}

\author{Seung-Hyun Song\orcidlink{0009-0001-4813-779X},~\IEEEmembership{Student~Member,~IEEE}, Dong-Hee Paek\orcidlink{0000-0003-0008-3726},~\IEEEmembership{~Member,~IEEE}, Woong-Chan Byun\orcidlink{0009-0003-0542-6860},~\IEEEmembership{Student~Member,~IEEE}, and Seung-Hyun~Kong\orcidlink{0000-0002-4753-1998},~\IEEEmembership{Senior~Member,~IEEE}
\thanks{This work was supported by the National Research Foundation of Korea(NRF) grant funded by the Korea government(MSIT) (No. 2021R1A2C3008370). \textit{(Seung-Hyun Song and Dong-Hee Paek are co-first authors.)} \textit{(Corresponding authors: Seung-Hyun Kong.)}}
\thanks{Seung-Hyun Song is with the Graduate School of Advanced Security Science and Technology, Korea Advanced Institute of Science and Technology, Daejeon, Korea, 34051 (e-mail: shyun@kaist.ac.kr)}
\thanks{Dong-Hee Paek, Woong-Chan Byun and Seung-Hyun Kong are with the CCS Graduate School of Mobility, Korea Advanced Institute of Science and Technology, Daejeon, Korea, 34051 (e-mail: donghee.paek@kaist.ac.kr; woongchan.byun@kaist.ac.kr; skong@kaist.ac.kr)}
}

\markboth{Preprint submitted to arXiv}%
{S.-H. Song \MakeLowercase{\textit{et al.}}: Stereo 4D Radar-Based 3D Object Detection}

\maketitle

\begin{abstract}
Four-dimensional (4D) Radar is a powerful sensing modality capable of detecting surrounding three-dimensional (3D) objects under diverse weather conditions and providing Doppler-based motion information. However, raw 4D Radar signals contain significant clutter from road surfaces, guardrails, and surrounding vehicles, along with multipath-induced ghost reflections and the receiver’s inherent noise floor. Consequently, preprocessing algorithms designed to remove such invalid measurements often make the Radar data excessively sparse. Moreover, the Doppler measurements provided by 4D Radar describe only the radial component of an object's velocity, limiting their ability to recover the full motion state.
In this paper, we introduce the stereo 4D Radar-based 3D object detection framework that exploits the geometric disparity between left and right Radars to estimate the absolute velocity of objects and to achieve more robust perception through the fusion of their complementary features. The effectiveness of the proposed framework is validated on our in-house stereo 4D Radar dataset, demonstrating performance gains of 8.82\% in $AP_{3D}$ and 9.0\% in $AP_{BEV}$ over state-of-the-art mono 4D Radar baselines. These results demonstrate that absolute velocity estimation combined with stereo geometry-aware feature fusion leads to substantial improvements in 3D object detection.
\end{abstract}


\section{Introduction}
\IEEEPARstart{R}{obust} perception that remains reliable under adverse weather conditions is essential for ensuring the safety and dependability of autonomous driving and intelligent transportation systems \cite{kitti,nuscenes,robustness}. In particular, accurate object detection serves as a core component of perception, enabling precise understanding of the driving environment as well as reliable decision-making and control \cite{Intro1, Intro2}. 

Recent advances in deep learning-based visual perception have driven autonomous driving research to focus primarily on LiDAR and camera sensors \cite{voxelnet, pointpillars, pvrcnn, monocular, bevfusion}. However, these optical sensors rely on visible and near-infrared light, whose measurements become unreliable under adverse weather conditions such as rain, fog, and snow. In addition, they cannot directly measure object velocity, and this combination ultimately results in reduced perception reliability \cite{adverse1,adverse2,adverse3}.
In contrast, Radar leverages long-wavelength millimeter-wave (mmWave) signals, providing robust perception performance even under harsh weather conditions. Moreover, Radar can directly measure the velocity of reflecting targets through the Doppler effect. Conventional automotive Radar has provided range, azimuth, Doppler velocity, and the power of each reflected signal. Recently, advances in high-resolution multiple-input multiple-output (MIMO) array technology have made it possible to measure elevation as well, giving rise to four-dimensional (4D) Radar systems. By providing three-dimensional (3D) spatial structure together with Doppler information, 4D Radar has emerged as a key sensing modality for robust perception in autonomous driving \cite{vod,tj4d,k-radar}.

\begin{figure*}
    \centering
    \includegraphics[width=0.9\linewidth]{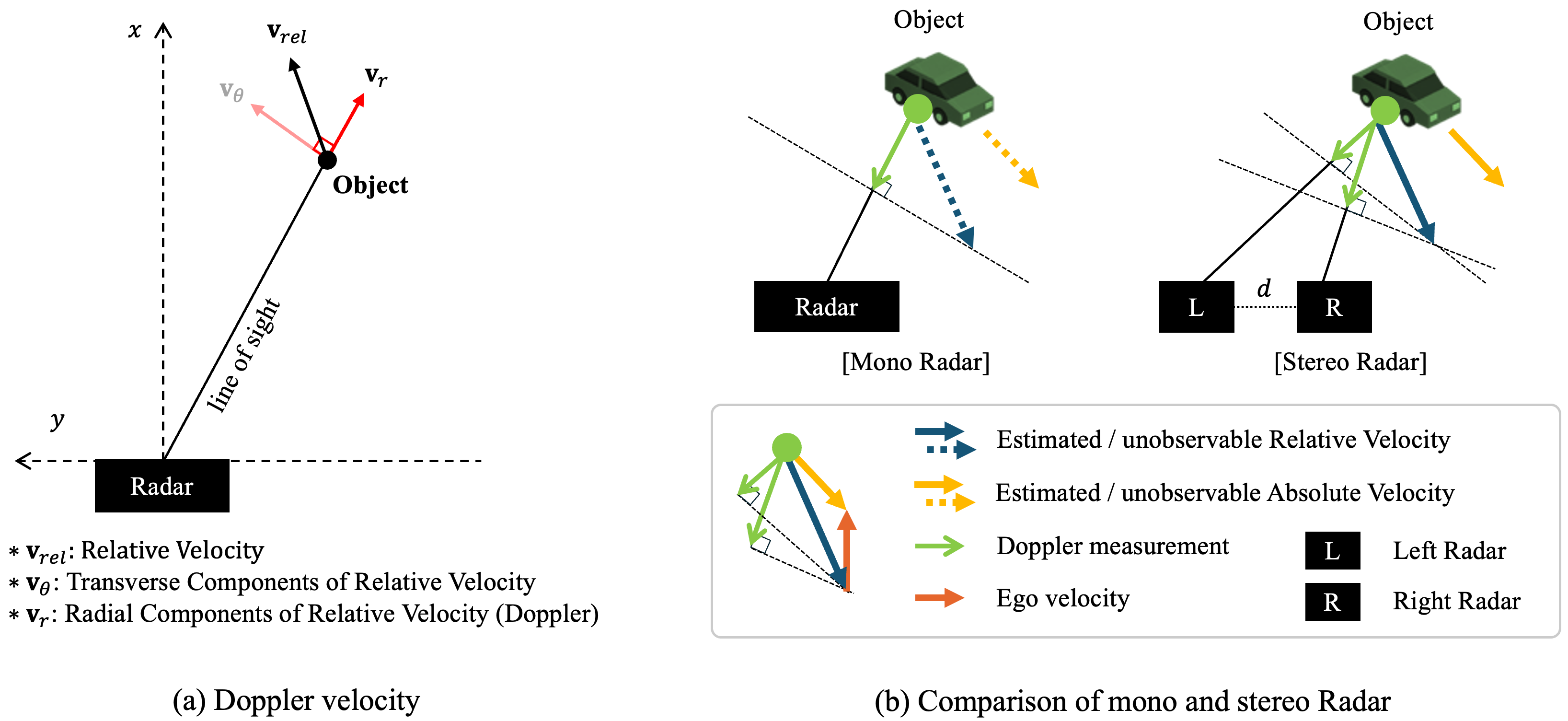}
    \caption{Stereo 4D Radar-based velocity estimation. (a) Illustration of Doppler velocity, which represents the radial component of an object's relative motion. (b) Stereo 4D Radar utilizes the Doppler measurements from the left and right Radars to estimate both the relative and absolute velocities.}
    \label{fig:1}
\end{figure*}

However, 4D Radar-based perception systems still face two critical challenges. 
First, as illustrated in Fig.~\ref{fig:1}-(a), the Doppler velocity measured by Radar contains only the radial component projected onto the line-of-sight (LoS) direction between the sensor and the target, and therefore cannot fully represent the object’s true motion. Prior studies have attempted to exploit this incomplete Doppler velocity to distinguish dynamic objects or to infer their motion direction for improved classification, and some approaches attempt to mitigate the limitations of incomplete Doppler measurements by incorporating ego-velocity information \cite{doppler1,doppler2,doppler3,doppler4,mfnet}.
Nevertheless, these methods remain limited by the fact that Doppler velocity is constrained to the LoS direction, making it difficult to recover the transverse components of object motion.

Second, 4D Radar measurements contain strong clutter from road surfaces, guardrails, and nearby vehicles, ghost reflections caused by multipath propagation, and receiver noise \cite{clutter1,clutter2}. Although preprocessing steps are applied to suppress these effects, they inevitably make the data excessively sparse and may remove valid returns along with the noise, limiting the reliability of subsequent perception modules \cite{preprocess1, preprocess2}.
To address these limitations, various approaches have been proposed, including encoding-based methods\cite{pointpillars,k-radar,rtnh+,rcfusion}, context-based methods\cite{rpfa,smurf,dadan}, and multi-representation-based methods\cite{mfnet,m3net,mvfan,smiformer,mufasa,maff}.
However, these approaches still rely on a mono 4D Radar sensor, whose limited field of view and inherently sparse point distribution reduce their ability to capture object shape and motion in real driving scenarios, resulting in performance that remains inferior to LiDAR- and camera-based perception systems.

To address these limitations, we propose a stereo 4D Radar-based 3D object detection framework that leverages geometric disparity between two Radars.
First, to address the incompleteness of Doppler velocity, we deploy two identical 4D Radars separated by a fixed baseline distance $d$, allowing the same object to be observed simultaneously from two distinct viewpoints. As illustrated in Fig.~\ref{fig:1}-(b), the geometric baseline between the Radars and the azimuth angles measured at each Radar allow the recovery of an object's absolute velocity from Doppler measurements. Building on this principle, we introduce a stereo 4D Radar-based absolute velocity estimation method and integrate it into a 3D object detection network.
In addition, to overcome the representational limitations from sparse Radar measurements, we introduce the Stereo Radar Pillar Feature Encoding (SR-PFE) and the Stereo Radar Bilateral Fusion Module (SR-BFM). These modules align, refine, and fuse the features from the left and right Radars, enabling effective integration of stereo 4D Radar information.
In this work, we design a framework that enhances incomplete motion information through stereo 4D Radar-based absolute velocity estimation and enriches sparse Radar representations through effective stereo fusion. Extensive experiments demonstrate that both components contribute substantially to improving 3D object detection performance.

In summary, the main contributions of this work are as follows:
\begin{itemize}
    \item To the best of our knowledge, we propose the first stereo 4D Radar framework for 3D object detection, effectively mitigating the sparsity and incomplete velocity information inherent to mono 4D Radar.
    \item We introduce a stereo 4D Radar-based absolute velocity estimation method that recovers the true motion of dynamic objects from Doppler measurements and demonstrate its contribution to improving detection performance.
    \item We develop a stereo 4D Radar-based 3D object detection network equipped with SR-PFE and SR-BFM, achieving improvements of 8.82\% in $AP_{3D}$ and 9.0\% in $AP_{BEV}$ over state-of-the-art mono 4D Radar baselines.
\end{itemize}

The remainder of this paper is structured as follows. Section II reviews related work on 4D Radar-based 3D object detection and Doppler utilization. Section III details the proposed absolute velocity estimation and detection network. Section IV presents the stereo 4D Radar dataset, experimental setup, and results. Section V concludes the paper and discusses future directions.

\begin{figure*}[t]
    \centering
    \includegraphics[width=1.0\linewidth]{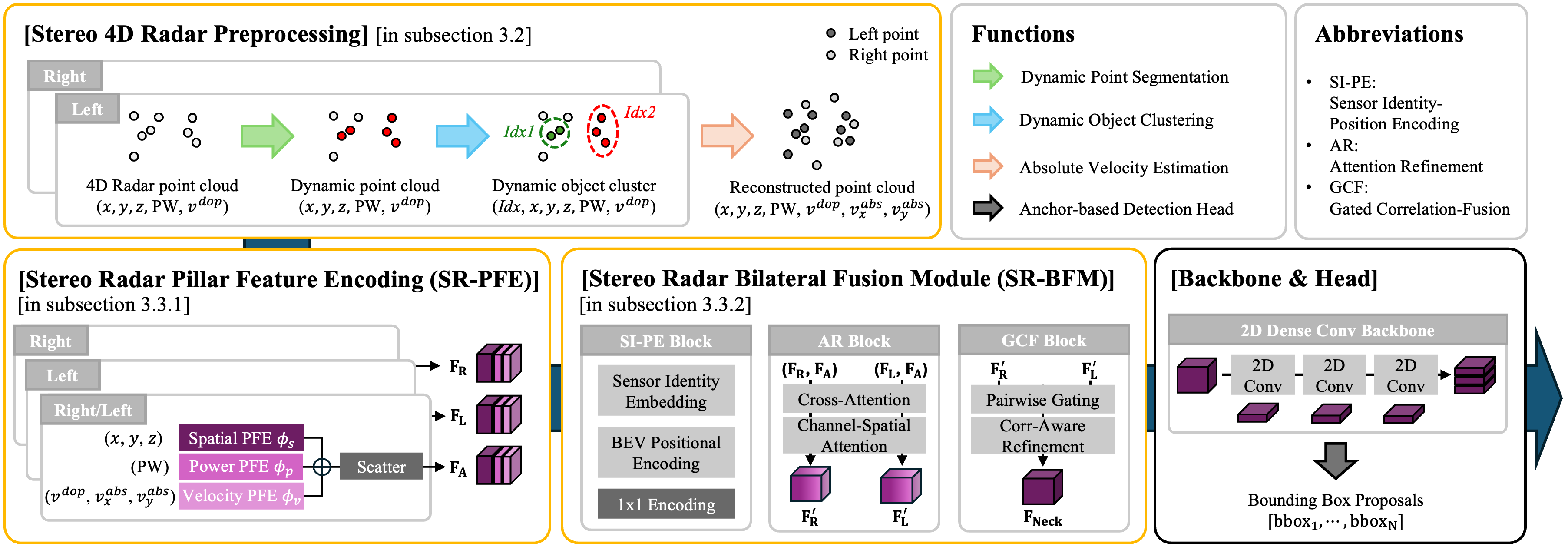}
    \caption{Overview of the proposed stereo 4D Radar-based 3D object detection pipeline. The framework consists of three key modules—Stereo 4D Radar Preprocessing, SR-PFE, and SR-BFM—along with a 2D backbone and an anchor-based detection head.}
    \label{fig:2}
\end{figure*}

\section{Related Work}
\subsection{3D Object Detection using 4D Radar}

To effectively extract geometric and semantic information from sparse 4D Radar points, various network architectures have been proposed. 
Early encoding-based approaches sought to preserve spatial structure. PointPillars \cite{pointpillars} forms pillar-wise representations via vertical quantization and learns BEV features using 2D CNNs. RTNH \cite{k-radar} and RTNH+ \cite{rtnh+} reconstruct Radar data into multi-dimensional tensors by separating the elevation dimension and independently processing spatial, velocity, and power channels to strengthen representation. Radar PillarNet \cite{rcfusion} adapts pillar encoding to Radar, generating pseudo-BEV images and enhancing BEV feature learning through separated and fused spatial, velocity, and power information.

To address missing local interactions and density cues in encoding, context-based methods have emerged. RPFA-Net \cite{rpfa} adds self-attention to model intra-pillar interactions and restore fine-grained representations. SMURF \cite{smurf} enhances pillar features with KDE-based density representations to reduce losses in sparse regions. DADAN \cite{dadan} leverages a density-aware module and Doppler-guided dynamic augmentation to achieve robust detection across diverse environments.

Recently, multi-representation methods have been explored to overcome the limitations of single-level representations and to maintain spatial and temporal consistency from multiple perspectives. RadarMFNet \cite{mfnet} aggregates features over consecutive frames and compensates point displacements using Doppler velocity, reducing motion distortion and yielding stable BEV representations. Radar M3-Net \cite{m3net} employs a multi-scale, multi-layer, and multi-frame architecture to enlarge the receptive field and hierarchically fuse features at different resolutions, improving detection of static and dynamic objects. MVFAN \cite{mvfan} extracts BEV and cylindrical representations in parallel and fuses their complementary spatial information through a multi-branch design. SMIFormer \cite{smiformer} models global interactions across BEV, front-view, and side-view representations using Transformer-based attention to integrate geometric relations among viewpoints. MUFASA \cite{mufasa} adaptively fuses range-angle, BEV, and cylindrical features via a spatial-awareness module for robust representation in diverse environments. MAFF-Net \cite{maff} enhances motion-aware BEV features by integrating auxiliary cues such as velocity, intensity, and Doppler maps through attention mechanisms.

Research on 4D Radar-based 3D perception has primarily centered on pillar representations for efficient BEV processing of sparse Radar points, while recent commercial developments have also explored lightweight real-time deployment.\footnote{For example, commercial systems demonstrated by DeepFusionAI.} Yet, even with advances across encoding, context, and multi-representation methods, mono 4D Radar with limited field of view and sparse point density continues to restrict perception performance.

\subsection{Use of Radar Doppler Information}
In addition to range- and angle-based spatial measurements, 4D Radar provides Doppler velocity, and numerous studies have explored leveraging this information for perception. RATRack \cite{doppler1} employs Doppler-
driven dynamic likelihood estimation to detect and track moving objects, while RSS-Net \cite{doppler2} incorporates Doppler as a class-discriminative feature for weakly-supervised semantic segmentation. CenterFusion \cite{doppler3} utilizes Doppler as an object-level motion hint to stabilize camera-Radar fusion, and RLNet \cite{doppler4} integrates Doppler into an adaptive fusion
module to better distinguish static and dynamic regions. In multi-frame approaches such as MFNet \cite{mfnet}, temporally aggregated Doppler is used to build velocity-consistent features for improved 3D object detection.

Meanwhile, most existing 4D Radar-based perception methods utilize the Doppler measurements in their incomplete form to support tasks. These approaches implicitly rely on Doppler as if it were sufficient motion information, even though it contains only the radial component. 

\section{Proposed Methods}
In this section, we first provide an overview of the overall architecture and operation flow of the proposed stereo 4D Radar-based 3D object detection network, and then describe each component in detail.

\subsection{Overview}
The overall architecture of the proposed stereo 4D Radar-based 3D object detection framework is illustrated in Fig.~\ref{fig:2}. The network consists of three key modules followed by a 2D convolutional backbone and an anchor-based detection head adapted from an open-source implementation,\footnote{Based on the publicly available OpenPCDet framework.} which generate the final object predictions.

In the Stereo 4D Radar Preprocessing stage, point clouds collected from the left and right Radar are used to estimate the ego-velocity, separate static background points from dynamic points, cluster dynamic objects, and estimate absolute velocity. Radar points augmented with absolute velocity information are then forwarded to the detection network. 
Next, SR-PFE encodes the spatial, power, and velocity information of each Radar point and converts them into BEV feature maps, producing separate feature representations for the left and right Radars.
SR-BFM consists of two components: an Attention Refinement (AR) block, which geometrically aligns and semantically refines the left and right features, and a Gated Correlation-Fusion (GCF) block, which generates a fused representation based on the reliability of the two Radar inputs.
The following sections provide a detailed description of each module and its role within the framework.

\begin{figure*}[t]
    \centering
    \includegraphics[width=1.0\linewidth]{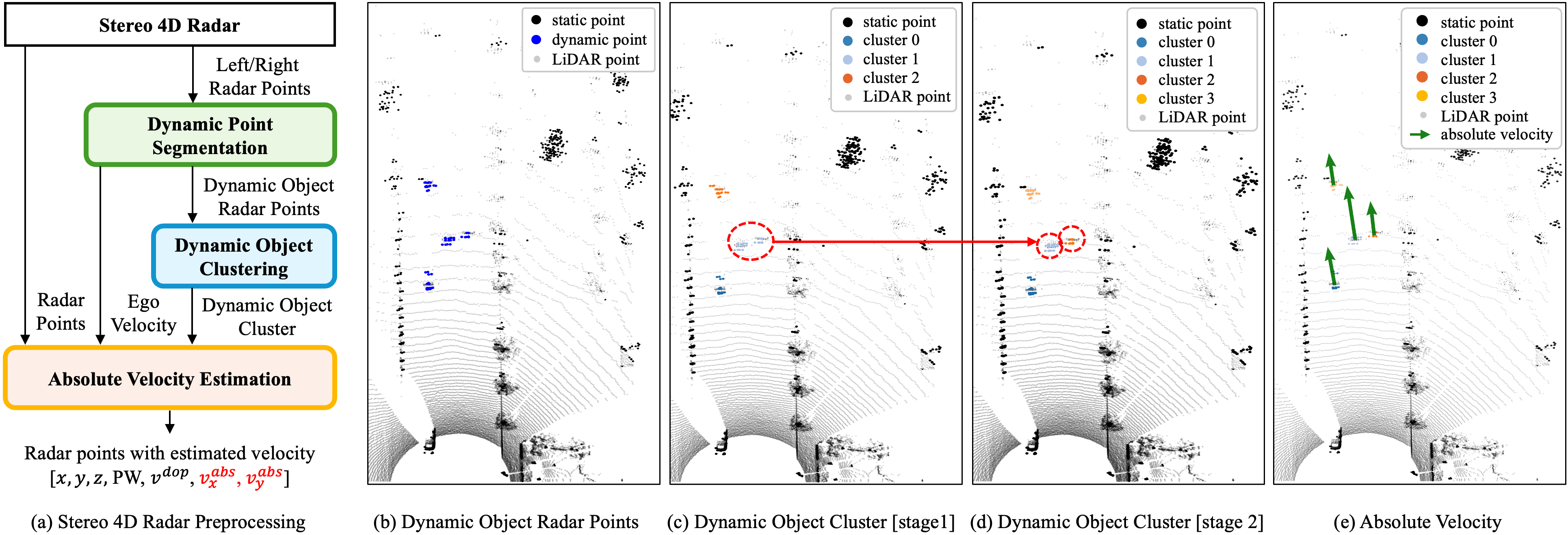}
    \caption{Stereo 4D Radar preprocessing pipeline and visualization of intermediate results.  
(a) Overall stereo 4D Radar preprocessing pipeline,  
(b) extraction of dynamic Radar points,  
(c) first-stage dynamic object clustering based on spatial proximity,  
(d) second-stage clustering using Doppler similarity,  
(e) absolute velocity estimation.  
The red-marked areas in (c) and (d) illustrate objects that spatial information alone fails to distinguish, whereas Doppler information enables clear separation.
}
    \label{fig:3}
\end{figure*}

\subsection{Stereo 4D Radar-based Velocity Estimation}
We propose a three-stage absolute velocity estimation framework tailored to the characteristics of stereo 4D Radar data.
First, dynamic points corresponding to moving objects are separated using Random Sample Consensus (RANSAC)-based Dynamic Point Segmentation \cite{ransac}. This process is performed independently for the left and right Radars, and the robustness of RANSAC to outliers enables stable segmentation even in the presence of noise or moving objects.
Second, dynamic object points from the left and right Radars are clustered using a two-step Density-Based Spatial Clustering of Applications with Noise (DBSCAN) procedure \cite{dbscan}. The first stage groups points based on spatial proximity, and the second stage refines the clusters using Doppler similarity, allowing accurate separation of adjacent objects with different motion patterns.
Third, absolute velocity is estimated using stereo-based ridge-regularized least squares (RLS). By combining Doppler measurements observed from two different viewpoints, stereo 4D Radar can recover the full velocity vector—including both radial and transverse components—rather than only the radial component available from a mono 4D Radar.

\subsubsection{Dynamic Point Segmentation}
In this stage, we apply a RANSAC-based method grounded in a 2D translation velocity model to (i) estimate the ego-velocity on the horizontal plane, denoted as $\hat{\mathbf{v}}_s$, using Doppler measurements, and (ii) separate points corresponding to dynamically moving objects to obtain a dynamic mask.

(i) Each Radar point is given by $\mathbf{p}_i\!=\!(x_i, y_i, z_i, \mathrm{PW}_i, v_i^{\mathrm{dop}})$,  
where $\mathrm{PW}_i$ denotes the reflection power and $v_i^{\mathrm{dop}}$ is the measured Doppler velocity.  
The LoS unit vector of each point, $\mathbf{h}_i$, is defined as:
\begin{equation}
    \mathbf{h}_i = \frac{[x_i, y_i, z_i]^\top}{\|[x_i, y_i, z_i]\|}, \quad
    \mathbf{h}_{i,xy} = [h_{i,x}, h_{i,y}]^\top,
\end{equation}
where $\mathbf{h}_{i,xy}$ denotes the projected unit vector on the horizontal plane $(x, y)$.
Based on the ground-plane prior that most object motion in road scenes occurs on the ground plane, 
the velocity estimation is restricted to the horizontal plane.
For static background points, the observed Doppler velocity is assumed to be caused solely by the ego-vehicle's
2D translational velocity $\mathbf{v}_s\!\in\!\mathbb{R}^2$.  
The expected Doppler velocity is therefore given by:
\begin{equation}
    \hat{v}_i^{\mathrm{dop}} = \mathbf{h}_{i,xy}^\top \mathbf{v}_s.
\end{equation}
The ego-velocity $\mathbf{v}_s$ is robustly estimated through RANSAC.  
At each iteration, a small subset of points is sampled to compute a tentative estimate of $\mathbf{v}_s$, 
and the residual errors between the estimated and measured Doppler values are evaluated over all points 
to form an inlier set $\mathcal{I}_{\mathrm{inliers}}$.  
Finally, a least-squares regression is applied to the inlier set to obtain the final ego-velocity estimate:
\begin{equation}
    \hat{\mathbf{v}}_s 
    = \min_{\mathbf{v}_s} \sum_{i \in \mathcal{I}_{\mathrm{inliers}}}
    \left( v_i^{\mathrm{dop}} - \mathbf{h}_{i,xy}^\top \mathbf{v}_s \right)^2.
    \label{eq:3}
\end{equation}
Points located close to the sensor or at the periphery of the field of view may exhibit large Doppler variations 
even for the same static object due to differences in observation angle.  
Therefore, inlier candidates are prioritized from distant points (beyond 15\,m) within the main viewing sector 
($\pm 40^\circ$) to ensure consistent and reliable estimation. The specific thresholds were empirically validated to provide the most reliable segmentation performance.

(ii) After estimating the ego-velocity $\hat{\mathbf{v}}_s$ from Eq.~\ref{eq:3},  
the expected Doppler velocity $\mathbf{h}_{i,xy}^\top \hat{\mathbf{v}}_s$ is computed for each point and 
compared with the measured Doppler value to obtain the residual $r_i$.  
The residual indicates how similar each point's velocity component is to the ego-motion:  
static background points exhibit small residuals, whereas independently moving objects show large values.
\begin{equation}
    r_i = \big|\, v_i^{\mathrm{dop}} - \mathbf{h}_{i,xy}^\top \hat{\mathbf{v}}_s \,\big|.
\end{equation}
Next, a threshold $\tau_{\text{base}}$ is determined based on the median of the residuals within 
$\mathcal{I}_{\mathrm{inliers}}$.  
The median is less sensitive to outliers than the mean, preventing distortion from noise or a small number 
of dynamic points and providing a stable measure of the central tendency of the residual distribution.  
Using this threshold, the dynamic mask $\mathcal{M}_{\text{dyn}}$ is obtained as follows:
\begin{equation}
    \tau_{\text{base}} = \mathrm{median}\big(\{\, r_i \mid i \in \mathcal{I}_{\mathrm{inliers}} \,\}\big),
\end{equation}
\begin{equation}
    \mathcal{M}_{\text{dyn}} = \{\, i \mid r_i > \tau_{\text{base}} \,\}.
\end{equation}
As shown in Fig.~\ref{fig:3}-(b), the proposed method clearly separates dynamic points (blue) such as vehicles 
from static background structures (black), including guardrails and trees, while effectively suppressing 
Doppler noise and outlier-induced errors.

\subsubsection{Dynamic Object Clustering}
After separating dynamic points, a two-step clustering process is applied to group points belonging to the same moving object.  
The input to this stage consists of the dynamic points identified in the previous step through the residual-based thresholding, represented by the dynamic mask $\mathcal{M}_{\mathrm{dyn}}$.

In the first stage, DBSCAN is applied to the dynamic points detected from the left and right Radars, 
grouping them into candidate clusters based on their 3D spatial proximity in $(x, y, z)$.  
Here, $\epsilon_{xyz}$ denotes the maximum inter-point distance that determines the spatial extent of a cluster, 
and $\text{minPts}_{xyz}$ specifies the minimum number of points required to form a valid cluster:
\begin{equation}
\mathrm{Cluster}_{xyz}\!
    =\!\mathrm{DBSCAN}(\{(x_i, y_i, z_i)\};\, \!\epsilon_{xyz},\,
    \!\text{minPts}_{xyz}).
\end{equation}
In the second stage, points within each spatial cluster are further grouped based on their Doppler velocities, 
allowing objects that are spatially adjacent but exhibit different motion patterns 
(e.g., two vehicles driving side by side) to be effectively separated:
\begin{equation}
\mathrm{Cluster}_{dop}
    \!=\!\mathrm{DBSCAN}(\{v_i^{\mathrm{dop}}\};\, \!\epsilon_{dop},\,\!\text{minPts}_{dop}).
\end{equation}
As illustrated in Fig.~\ref{fig:3}-(c) and \ref{fig:3}-(d), objects that cannot be distinguished using spatial information alone 
become clearly separable through the second-stage Doppler-based clustering.

\subsubsection{Absolute Velocity Estimation}
The absolute velocity estimation problem is formulated as a regularized RLS model using the Doppler measurements from the two sensors and the geometric relationship between each Radar and the target.

For a dynamic cluster consisting of $N_p$ Radar points, let $\mathbf{h}_i\!\in\!\mathbb{R}^2$ denote the LoS unit vector for point $i$,  
and let $\mathrm{d}_i$ represent the Doppler measurement corrected by the estimated ego-velocity.  
This can be written as:
\begin{equation}
    \mathrm{d}_i = v_i^{dop} + \mathbf{h}_i^\top \hat{\mathbf{v}}_{s}.
\end{equation}
This relationship expands the observation model at each point $i$ into $N_p$ independent constraints,  
which can be stacked to form the following linear system:
\begin{equation}
    \mathbf{d} = \mathbf{H}\mathbf{v},
\end{equation}
where $\mathbf{d}\!=\![\mathrm{d}_1, \mathrm{d}_2, \dots, \mathrm{d}_{N_p}]^\top\!\in\!\mathbb{R}^{N_p}$  
is the vector of ego-corrected Doppler measurements,  
$\mathbf{H}\!=\![\mathbf{h}_1^\top; \dots; \mathbf{h}_{N_p}^\top]$ is the matrix of LoS unit vectors,  
and $\mathbf{v} \in \mathbb{R}^2$ denotes the object's absolute (ground-truth) velocity.
The velocity estimation can then be formulated as a regularized least-squares problem:
\begin{equation}
    \hat{\mathbf{v}}
    = \arg\min_{\tilde{\mathbf{v}}}
    \|\mathbf{H}\tilde{\mathbf{v}} - \mathbf{d}\|_2^2
    + \lambda \|\tilde{\mathbf{v}}\|_2^2,
\end{equation}
where $\lambda$ is a regularization parameter that suppresses numerical instability when the LoS vectors become nearly parallel, 
$\tilde{\mathbf{v}}$ is the optimization variable, and $\hat{\mathbf{v}}$ denotes the estimated velocity obtained from the regularized least-squares solution.

The proposed stereo 4D Radar-based velocity estimation is significant in that it enables direct recovery of absolute velocity components that cannot be reconstructed using a mono sensor.  
By incorporating a regularization term, the method alleviates numerical instability that may arise when the LoS vectors are nearly parallel or when measurements are corrupted by noise, allowing robust estimation even under insufficient geometric constraints.  
The effectiveness and advantages of the proposed absolute velocity estimation are demonstrated in the experimental section through comparisons with mono 4D Radar-based velocity estimation methods.

\subsection{Stereo 4D Radar-based 3D Object Detection Network}
In this section, we focus on the key components designed to effectively process stereo 4D Radar inputs: the Stereo Radar Pillar Feature Encoder (SR-PFE) and the Stereo Radar Bilateral Fusion Module (SR-BFM).
SR-PFE encodes the various attributes of Radar measurements to generate rich BEV representations, while SR-BFM leverages stereo geometric information by aligning, refining, and fusing the left and right Radar features.

\begin{figure*}[t]
    \centering
    \includegraphics[width=1.0\linewidth]{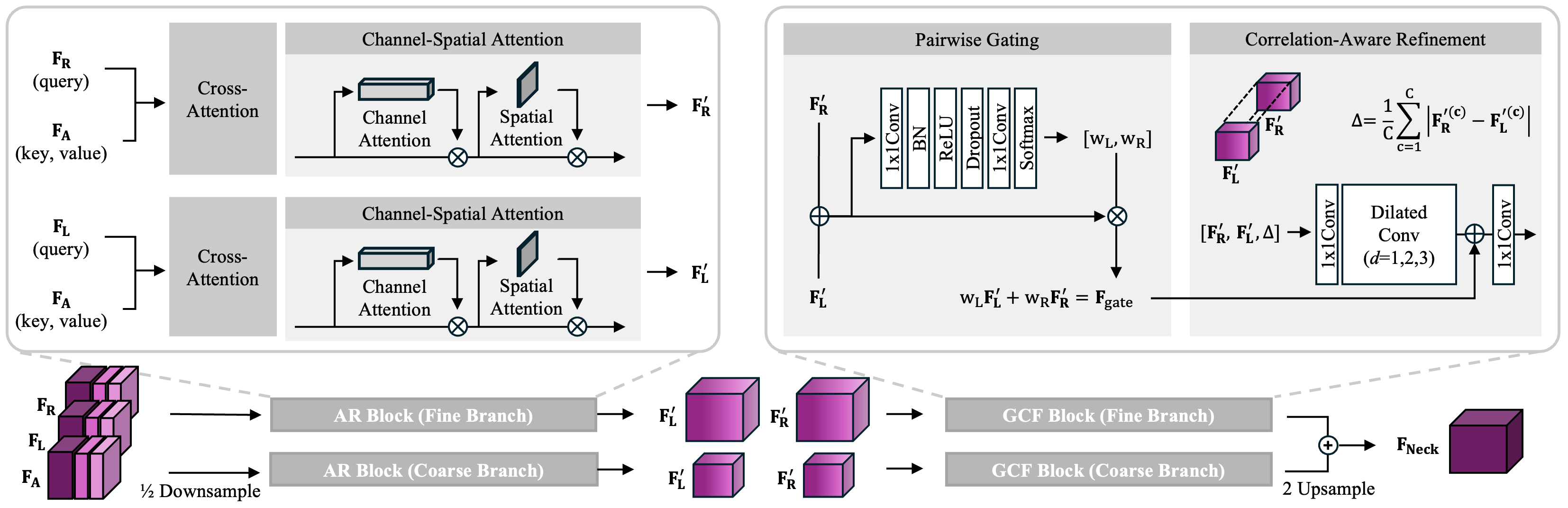}
    \caption{Overview of the SR-BFM architecture. SR-BFM comprises an Attention Refinement (AR) block, which aligns and enhances stereo features, and a Gated Correlation-Fusion (GCF) block, which integrates them through correlation-aware refinement and pairwise gating.}
    \label{fig:4}
\end{figure*}

\subsubsection{Stereo Radar Pillar Feature Encoding (SR-PFE)}
The proposed SR-PFE extends the encoding structure of Radar PillarNet\cite{rcfusion}. It first partitions the input Radar point cloud into spatial pillars and then encodes the points within each pillar to generate BEV feature maps.

We use three types of Radar inputs: the left sensor (L), the right sensor (R), and an aggregated input (A) obtained by concatenating the left and right Radar point sets. 
For each input $k\!\in\!\{\mathrm{L, R, A}\}$, 
let $\mathcal{P}^k\!=\!\{\mathbf{p}^k_j\}_{j=1}^{M}$ denote the set of $M$ points contained within a pillar.  
From these points, three types of feature vectors are constructed: a spatial feature vector $\mathbf{f}^k_{s,j}$ using $\{x, y, z\}$, 
a reflection feature vector $\mathbf{f}^k_{r,j}$ using $\{\mathrm{PW}\}$, 
and a velocity feature vector $\mathbf{f}^k_{v,j}$ using  
$\{v^{dop}, v^{abs}_x, v^{abs}_y\}$.  
These feature vectors are passed into three encoders:  
a spatial encoder $\phi_s(\cdot)$, a power encoder $\phi_r(\cdot)$, 
and a velocity encoder $\phi_v(\cdot)$.
Each encoder produces a point-wise embedding, which is concatenated channel-wise, 
followed by max pooling within each pillar and scattering onto the BEV plane to obtain a sensor-specific feature map $\mathbf{F}_k$. 
The entire process can be summarized as:
\begin{equation}
\mathbf{F}_k\!=
\!\mathrm{Scatter}\!\left(\!
\max_j
\left[\!
\,
\big[
\phi_s(\mathbf{f}^k_{s,j})
\,;\,\!
\phi_r(\mathbf{f}^k_{r,j})
\,;\,\!
\phi_v(\mathbf{f}^k_{v,j})
\big]
\right]\!
\right).
\end{equation}
The generated feature maps 
$\mathbf{F}_{\mathrm{L}}$, $\mathbf{F}_{\mathrm{R}}$, and $\mathbf{F}_{\mathrm{A}}$ 
represent the BEV features extracted from the left Radar, right Radar, and aggregated input, respectively, and are used as inputs to the subsequent SR-BFM module.

\subsubsection{Stereo Radar Bilateral Fusion Module (SR-BFM)}
To integrate the complementary features of stereo 4D Radar, we propose the SR-BFM. SR-BFM consists of three sequential components — the Sensor Identity-Positional Encoding (SI-PE) block, the Attention Refinement (AR) block, and the Gated Correlation-Fusion (GCF) block — which progressively align and fuse the left-right Radar features.
Each block plays a distinct role in sensor identification, feature alignment, and feature fusion, and their effectiveness is quantitatively validated through ablation studies in Section~\ref{sec:ablation}.

\paragraph{\textbf{SI-PE Block}}
Given the Radar feature maps extracted from the left, right, and aggregated encoders, the SI-PE Block enriches the representation by injecting sensor identity and positional information. First, a learnable Sensor Identity Embedding is applied to explicitly distinguish the inputs from each sensor. Then, a Fourier Positional Encoding module adds global spatial context over the BEV grid. Finally, a 1×1 convolution is used to align the feature dimensions, producing representations that are well suited for the subsequent fusion stages.

\paragraph{\textbf{AR Block}}
The AR Block is responsible for aligning the left and right Radar features by mitigating spatial inconsistencies between sensors and producing representations that are interpretable within a shared embedding space. The block consists of two parallel branches, where each branch takes the features from one sensor as queries and applies local cross attention against the aggregated feature map $\mathbf{F_A}$ used as keys and values. Formally, for the right- and left-Radar features $\mathbf{F_R}$ and $\mathbf{F_L}$, the local cross attention operations are defined as:
\begin{equation}
\tilde{\mathbf{F}}_\mathbf{R}\!=\!\mathrm{CrossAttn}(\mathbf{F_R}, \mathbf{F_A}),
\tilde{\mathbf{F}}_\mathbf{L}\!=\!\mathrm{CrossAttn}(\mathbf{F_L}, \mathbf{F_A}),
\end{equation}
where $\mathrm{CrossAttn}(\cdot,\cdot)$ denotes a localized cross-attention operator that computes query-key similarity and weighted aggregation within a restricted spatial window.
This process effectively aligns the left-right sensor observations to the unified coordinate frame provided by $\mathbf{F_A}$, enabling consistent fusion in subsequent stages.

After local geometric alignment, the AR Block further enhances feature quality by learning to reweight the channel and spatial dimensions, thereby amplifying semantically salient structures (e.g., vehicles, pedestrians) while suppressing background noise. Let $\mathrm{CSAttn}(\cdot)$ denote the channel-spatial attention function. The refined stereo features are then obtained as:
\begin{equation}
\mathbf{F'_R}\!=\!\mathrm{CSAttn}(\tilde{\mathbf{F}}_\mathbf{R}),
\qquad
\mathbf{F'_L}\!=\!\mathrm{CSAttn}(\tilde{\mathbf{F}}_\mathbf{L}),
\end{equation}

Through the combination of local cross attention-based geometric alignment and channel-spatial importance reweighting, the AR Block produces semantically refined and geometrically consistent stereo features, $\mathbf{F'_R}$ and $\mathbf{F'_L}$. These refined representations are then passed to the GCF Block, which integrates complementary stereo information for final fusion.

\begin{table*}[t!]
\setlength{\tabcolsep}{12pt}
\centering
\caption{Comparison of 3D object detection performance with different Radar input configurations, examining the effects of stereo input and absolute velocity incorporation.}
\label{tab:object_detection_result}
\begin{tabular}{cl|cc|cccc}
\hline \hline
\multicolumn{2}{c|}{} & \multicolumn{2}{c|}{} & \multicolumn{4}{c}{Stereo} \\ \cline{5-8} 
\multicolumn{2}{c|}{} & \multicolumn{2}{c|}{\multirow{-2}{*}{Mono}} & \multicolumn{2}{c|}{without $(v^{abs}_{x},v^{abs}_{y})$} & \multicolumn{2}{c}{with $(v^{abs}_{x},v^{abs}_{y})$} \\ \cline{3-8} 
\multicolumn{2}{c|}{\multirow{-3}{*}{Network}} & $AP_{3D}$ & $AP_{BEV}$ & $AP_{3D}$ & \multicolumn{1}{c|}{$AP_{BEV}$} & $AP_{3D}$ & $AP_{BEV}$ \\ \hline \hline
\multicolumn{2}{c|}{RTNH\cite{k-radar}}  & 49.10 & 49.82 & 56.72 & \multicolumn{1}{c|}{57.37} & 57.03 & 57.71 \\ \hline
\multicolumn{2}{c|}{RTNH+\cite{rtnh+}} & 50.10 & 50.73 & 57.14 & \multicolumn{1}{c|}{57.83} & 57.29 & 57.93 \\ \hline
\multicolumn{2}{c|}{RadarPillarNet\cite{rcfusion}} & 48.02 & 49.52 & 57.23 & \multicolumn{1}{c|}{58.25} & 58.01 & 58.95 \\ \hline
\multicolumn{2}{c|}{SMURF\cite{smurf}} & 39.93 & 41.84 & 50.37 & \multicolumn{1}{c|}{50.89} & 50.46 & 51.05 \\ \hline
\multicolumn{2}{c|}{RPFANet\cite{rpfa}} & 41.07 & 41.74 & 48.89 & \multicolumn{1}{c|}{49.92} & 48.78 & 56.40 \\ \hline
\multicolumn{2}{c|}{RadarMFNet\cite{mfnet}} & 47.77 & 48.86 & 49.71 & \multicolumn{1}{c|}{50.48} & 50.20 & 57.93 \\ \hline\multicolumn{2}{c|}{MVFAN\cite{mvfan}} & 49.22 & 50.33 & 57.27 & \multicolumn{1}{c|}{58.24} & 57.34 & 58.35 \\ \hline
\multicolumn{2}{c|}{DADAN\cite{dadan}} & 49.99 & 50.69 & 57.03 & \multicolumn{1}{c|}{57.93} & 57.66 & 58.51 \\ \hline
\multicolumn{2}{c|}{\textbf{Proposed}} & - & - & 58.56 & \multicolumn{1}{c|}{59.56} & \textbf{58.92} & \textbf{59.73}
\\ \hline \hline
\end{tabular} 
\end{table*}

\paragraph{\textbf{GCF Block}}
The GCF block consists of two main components:
(i) a pairwise gating module that adaptively selects reliable features from each sensor, and
(ii) a correlation-aware refinement module that further aligns and enhances the fused representation.

In the pairwise gating stage, the BEV-wise confidence weights 
$\mathbf{w}\!=\![\mathrm{w_R},\, \mathrm{w_L}]\!\in \!\mathbb{R}^{2 \times H \times W}$ 
are estimated by jointly considering the refined left-right features 
$\mathbf{F}'_{\mathbf{R}}$ and $\mathbf{F}'_{\mathbf{L}}$. 
As illustrated in Fig.~\ref{fig:4}, the two feature maps are concatenated along the channel dimension and passed through a lightweight head (1×1 Conv-BN-ReLU-Dropout-1×1 Conv), followed by a softmax normalization that enforces 
$\mathrm{w_R}\!+\!\mathrm{w_L}\!=\!1$ at each BEV location:
\begin{equation}
\mathbf{w}\!=\!\mathrm{Softmax}\big(\Gamma([\mathbf{F'_R},\, \mathbf{F'_L}])\big),
\qquad
\mathbf{w}\!=\![\mathrm{w_R},\, \mathrm{w_L}],
\end{equation}
where $\Gamma(\cdot)$ denotes a lightweight prediction module composed of a 
$1{\times}1$ convolution, normalization, non-linear activation, and dropout.
The initial fused feature $\mathbf{F}_{\text{gate}}$ is then computed as:
\begin{equation}
\mathbf{F}_{\textbf{gate}}\!=\!\mathrm{w_R} \odot \mathbf{F'_R}\!+\!\mathrm{w_L} \odot \mathbf{F'_L}.
\end{equation}
This weighted combination allows the network to rely more heavily on the viewpoint 
with higher confidence, ensuring robust fusion even when one sensor suffers from 
occlusion, weak reflections, or partial signal degradation.

In the correlation-aware refinement stage, the three inputs 
$\mathbf{F}'_{\mathbf{R}}$, $\mathbf{F}'_{\mathbf{L}}$, and  
$\Delta\!=\!\frac{1}{C}\sum_{c=1}^{C} |\mathbf{F}'^{(c)}_{\mathbf{R}} - \mathbf{F}'^{(c)}_{\mathbf{L}}|$ 
are concatenated and processed in parallel by dilated convolutions 
$\mathrm{Conv}_{d}(\cdot)$ with dilation rates $d \in \{1,2,3\}$. 
These operations highlight and correct regions of inconsistency between the two sensors, 
reducing geometric discrepancies:
\begin{equation}
\mathbf{F}_{\textbf{ref}}\!=\!\mathrm{Conv}_{1\times1}\!\left([\mathrm{Conv}_{k}(\mathbf{X})]_{k=1}^{3}\right),
\end{equation}

\begin{equation}
\mathbf{X}\!
=\!\mathrm{Concat}[\mathbf{F'_R},\, \mathbf{F'_L},\, \Delta].
\end{equation}
The refined feature $\mathbf{F}_{\text{ref}}$ is then concatenated with the 
gated fusion output $\mathbf{F}_{\text{gate}}$ along the channel dimension, 
and a $1{\times}1$ convolution produces the final fused feature map:
\begin{equation}
\mathbf{F_{Neck}}\!=\! 
\mathrm{Conv_{1\times1}}\!\left(
\mathrm{Concat}[\mathbf{F_{ref}},\, \mathbf{F_{gate}}]
\right).
\end{equation}

As a result, the GCF Block combines confidence-based gating with correlation-aware refinement to correct left-right discrepancies and produce a spatially consistent, semantically enhanced fused representation $\mathbf{F}_{\textbf{Neck}}$
This fused feature map is subsequently fed into the backbone and detection head to generate the final 3D object detection outputs.

\section{Experiments}
In this section, we evaluate the performance of the proposed stereo 4D Radar-based 3D object detection framework. To this end, we construct a new stereo 4D Radar dataset that captures the same scenes from two spatially separated sensors. All experiments are conducted using this dataset, enabling a comprehensive assessment of both absolute velocity estimation and the resulting improvements in 3D object detection performance.

\subsection{Dataset}
For this study, we constructed a multimodal dataset featuring stereo 4D Radars as the primary sensing modality, complemented by a LiDAR sensor for annotation and multiple forward-facing cameras for visualization.

The dataset comprises 37 sequences (22K frames) recorded in clear daytime driving scenarios, with about 66K 3D bounding boxes annotated for the \emph{Sedan} class. Compared to existing public 4D Radar datasets, the scale of our dataset is sufficiently large for quantitative evaluation and generalization analysis \cite{astyx, radial, vod, tj4d}. The structure and detailed specifications of the dataset are provided in the Appendix~\ref{sec:appendix}.

\subsection{Experimental Setup and Metric}
The proposed model is implemented using the PyTorch 1.12 framework and trained on an Ubuntu~22.04 system equipped with an NVIDIA RTX~4070~Ti GPU (12\,GB). Training is performed for 30 epochs with an initial learning rate of 0.001 and a batch size of 8. The Region of Interest (RoI) for Radar points is set to $x \in [0, 72]$\,m, $y \in [-16, 16]$\,m, and $z \in [-2, 7.6]$\,m.

For dynamic object clustering, the spatial proximity threshold is set to $\epsilon_{xyz}\!=\!2.5$\,m 
and the Doppler similarity threshold to 
$\epsilon_{dop}\!=\!0.5$\,m/s, enabling both spatial separation of vehicles and fine-grained grouping based on velocity differences. 
We use $\text{minPts}_{xyz}\!=\!6$ for the first-stage spatial clustering to ensure sufficient point density, and $\text{minPts}_{dop}\!=\!2$ in the second stage to separate objects with narrow velocity distributions (e.g., distant vehicles).

For 3D object detection evaluation, we follow the K-Radar benchmark\cite{k-radar} protocol and adopt Average Precision ($AP$) as the performance metric. Specifically, we report $AP_{3D}$ and $AP_{BEV}$ for the \emph{Sedan} class at an intersection-over-union (IoU) threshold of 0.3.

\begin{figure*}[t!]
    \centerline{\includegraphics[width=0.9\linewidth]{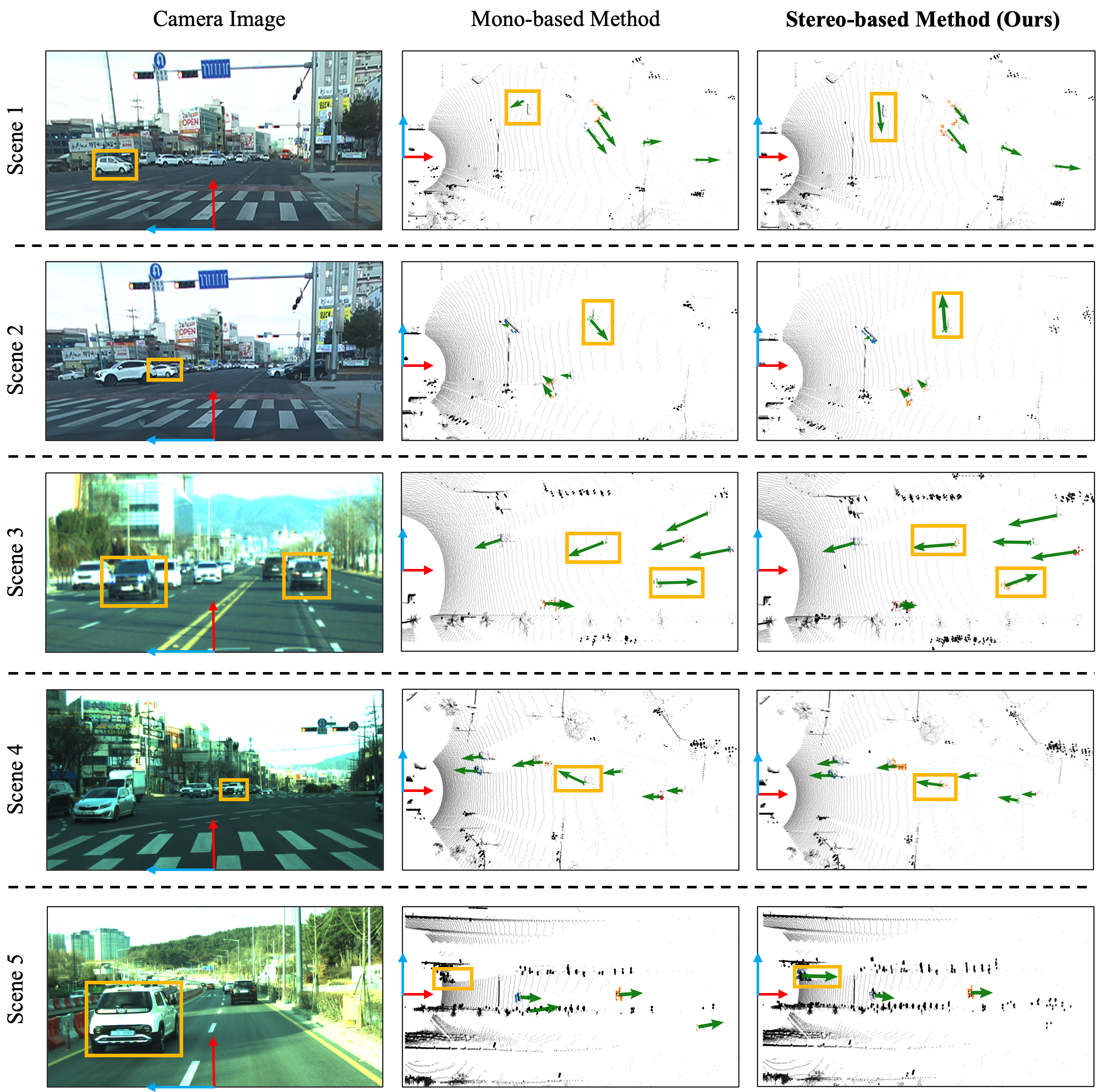}}
    \caption{
        Qualitative comparison of velocity estimation between mono-based and stereo-based methods. Red and blue arrows indicate the ego-vehicle’s local X- and Y-axes, respectively, while green arrows represent the absolute velocity vectors of surrounding objects. Yellow boxes highlight notable dynamic objects for comparison.
        }
    \label{fig:5}
\end{figure*}

\subsection{Quantitative Analysis of Mono vs. Stereo 4D Radar Configurations}
Table~\ref{tab:object_detection_result} presents a quantitative comparison of 3D object detection performance across various Radar-based networks under different input configurations.
This evaluation aims to verify that the proposed stereo 4D Radar framework effectively mitigates the two fundamental limitations highlighted in the Introduction:
(1) the sparsity of mono 4D Radar measurements after preprocessing, and
(2) the incomplete and noisy Doppler velocity observations that lead to loss of reliable motion information.

First, we compare the performance of mono-Radar input with that of stereo 4D Radar input.
Across all networks, introducing stereo input consistently improves both $AP_{3D}$ and $AP_{BEV}$.
For example, in RadarPillarNet, $AP_{3D}$ increases from 48.02 to 57.23 and $AP_{BEV}$ rises from 49.52 to 58.25.
Similar trends appear in other networks such as RTNH\cite{k-radar} and RadarMFNet\cite{mfnet}, demonstrating that combining spatially complementary observations from the left and right Radars effectively alleviates the extreme sparsity of mono 4D Radar measurements and enables the models to learn richer geometric representations.

Next, we compare stereo input using only Doppler measurements with stereo input augmented by the estimated absolute velocities.
In most networks, adding absolute velocity information further improves both $AP_{3D}$ and $AP_{BEV}$.
This indicates that the motion information recovered through stereo-based velocity estimation can be effectively utilized by existing detectors, enabling more accurate localization and motion interpretation of dynamic objects.

The proposed network achieves the highest performance with $AP_{3D}$ of 58.92 and $AP_{BEV}$ of 59.73, outperforming all existing 4D Radar-based detectors.
These results confirm that our framework successfully integrates geometric complementarity from stereo 4D Radar with accurate absolute velocity estimation, thereby overcoming both the sparsity and incomplete Doppler limitations inherent to mono 4D Radar systems and significantly boosting 3D object detection performance.

\subsection{Absolute Velocity Estimation}
In this section, we demonstrate the effectiveness of the proposed stereo 4D Radar-based velocity estimation in recovering the true motion of dynamic objects. To this end, we provide a qualitative comparison against conventional mono-Radar-based velocity estimation.
Fig.~\ref{fig:5} presents representative driving scenes, showing the camera image (left), the mono-based estimation results (middle), and the proposed stereo 4D Radar-based estimation results (right).

Scene 1 and Scene 2 compare absolute velocity estimation for vehicles moving laterally. The mono-based method struggles to capture transverse motion, often producing distorted velocity vectors. In contrast, the stereo-based method leverages Doppler measurements from both left and right sensors, effectively compensating for missing motion components and recovering the true motion of the target object.

Scene 3 and Scene 4 highlight driving situations where vehicles exhibit complex motion patterns that cannot be reliably inferred from a mono 4D Radar viewpoint. In Scene 3, for example, the oncoming vehicle in the left lane (left yellow box) maintaining a straight trajectory, and the ego-lane vehicle (right yellow box) initiating a lane change to the left. The mono-based method misinterprets their motion directions due to the lack of complete velocity, whereas the stereo-based estimation clearly separates radial and transverse components, yielding accurate motion vectors.

Finally, Scene 5 highlights a case where the stereo configuration improves motion perception compared to the mono-Radar. The mono-based method fails to correctly classify a dynamic object, whereas the stereo approach utilizes the geometric complementarity between the two viewpoints to recover missing motion information and reliably identify the object as dynamic.

These results show that the proposed stereo 4D Radar-based velocity estimation robustly compensates for missing or incomplete motion information in various driving scenarios.

\begin{table*}[]
\centering
\small
\caption{Impact of SR-PFE and SR-BFM on Stereo 4D Radar Detection.
$N_a$, $N_b$, $N_c$, and $N_d$ represent variants where the SI-PE Block, AR Block, GCF Block, and SR-PFE are removed, respectively.
}
\vspace{0.5em}
\label{tab:2}
\begin{tabular}{c|c|ccc|cc}
\hline \hline
\multirow{2}{*}{Network} & \multirow{2}{*}{SR-PFE} & \multicolumn{3}{c|}{SR-BFM} & \multirow{2}{*}{$AP_{BEV}$} & \multirow{2}{*}{$AP_{3D}$} \\ \cline{3-5}
 &  & SI-PE Block & AR Block & GCF Block &  &  \\ \hline
$N_a$ & \checkmark &  & \checkmark & \checkmark & 58.87 & 57.91 \\
$N_b$ & \checkmark & \checkmark &  & \checkmark & 58.76 & 57.83 \\
$N_c$ & \checkmark & \checkmark & \checkmark &  & 58.02 & 56.72 \\
$N_d$ &  & \checkmark & \checkmark & \checkmark & 59.10 & 58.01 \\
\hline
\textbf{ours} & \checkmark & \checkmark & \checkmark & \checkmark & \textbf{59.73} & \textbf{58.92} \\ \hline \hline
\end{tabular}
\end{table*}

\subsection{Ablation Study}
\label{sec:ablation}
This section quantitatively analyzes the contribution of each component within the proposed SR-BFM and examines how the module’s architectural design influences the overall 3D object detection performance.

\subsubsection{Contribution Analysis of SR-PFE and SR-BFM}

Table~\ref{tab:2} summarizes the contribution of each component within the proposed SR-BFM—namely the SI-PE Block, AR Block, and GCF Block—as well as the effect of removing SR-PFE.

Removing the GCF Block ($N_c$) yields the largest degradation, with $AP_{BEV}=58.02$ and $AP_{3D}=56.72$, indicating that correlation-aware refinement and gating are central to effective stereo fusion. 
Eliminating the SI-PE Block ($N_a$) reduces performance to $AP_{BEV}=58.87$ and $AP_{3D}=57.91$, showing that sensor identity embeddings provide a stable reference frame for stereo alignment. 
Removing the AR Block ($N_b$) lowers accuracy to $AP_{BEV}=58.76$ and $AP_{3D}=57.83$, confirming that the cross-attention-based alignment mitigates left-right geometric inconsistencies. 
Excluding SR-PFE ($N_d$) also results in a noticeable drop ($AP_{BEV}=59.10$, $AP_{3D}=58.01$), demonstrating its role in producing discriminative BEV features prior to stereo fusion.

The full model (ours) achieves the highest performance, with $AP_{BEV}=59.73$ and $AP_{3D}=58.92$, verifying that SR-PFE and all SR-BFM components jointly enhance stereo 4D Radar detection.

\begin{table}[]
\centering
\small
\caption{
Performance comparison of the SR-BFM sub-components (Coarse Branch and Dilated Convolution). $N_a$ denotes the variant without the Coarse Branch, and $N_b$ denotes the variant without Dilated Convolution.
}
\vspace{0.5em}
\setlength{\tabcolsep}{2.7pt} 
\label{tab:3}
\begin{tabular}{c|c|c|cc}
\hline \hline
\multirow{2}{*}{Network} & \multirow{2}{*}{Coarse Branch} & \multirow{2}{*}{Dilated Convolution} & \multirow{2}{*}{$AP_{BEV}$} & \multirow{2}{*}{$AP_{3D}$} \\
 &  &  &  &  \\ \hline 
$N_a$ & \checkmark &  & 59.39 & 58.34 \\
$N_b$ &  & \checkmark & 59.36 & 58.27 \\
\hline
\textbf{ours} & \checkmark & \checkmark & \textbf{59.73} & \textbf{58.92} \\ \hline \hline
\end{tabular}
\end{table}

\subsubsection{Effectiveness of Detailed Components in the GCF Block}
Table~\ref{tab:3} reports the effectiveness of the detailed components within the GCF Block, namely the Coarse Branch and the Dilated Convolution. Removing the Coarse Branch ($N_a$) results in a noticeable performance drop, with $AP_{BEV} = 59.39$ and $AP_{3D} = 58.34$. 
This suggests that capturing low-resolution global context is essential for maintaining wide-range alignment consistency during stereo feature fusion. When the Dilated Convolution 
module is removed ($N_b$), the performance declines further to $AP_{BEV} = 59.36$ and $AP_{3D} = 58.27$. This demonstrates that dilated receptive fields effectively correct geometric discrepancies. 
The full model (ours), which incorporates both components, achieves the highest performance with $AP_{BEV} = 59.73$ and $AP_{3D} = 58.92$, confirming that the multi-scale alignment 
and correlation-aware refinement jointly contribute to improving stereo fusion quality.

\section{Conclusion}
In this paper, we addressed two fundamental limitations of 4D Radar perception—%
the severe sparsity of point clouds after preprocessing and the incomplete motion information caused by the radial nature of Doppler measurements. To overcome these challenges, we introduced a new stereo 4D Radar-based 3D object detection framework that restores the absolute motion of dynamic objects and effectively fuses the complementary observations from left and right sensors. The proposed design integrates stereo-based absolute velocity estimation with cross-attention-driven alignment and correlation-aware refinement, enabling robust geometric consistency and enhanced representation quality beyond the capabilities of mono 4D Radar methods.

Extensive experiments on our stereo 4D Radar dataset demonstrate that the proposed framework outperforms existing state-of-the-art 4D Radar-based 3D object detection. In future work, we plan to extend the proposed framework toward tracking and multi-object motion reasoning using stereo 4D Radar, aiming to further enhance the completeness and reliability of Radar-based perception.

\appendix
\section{Stereo 4D Radar Dataset}
\label{sec:appendix}
\subsection{Sensor Specification}
\begin{figure}
    \centering
    \includegraphics[width=0.9\linewidth]{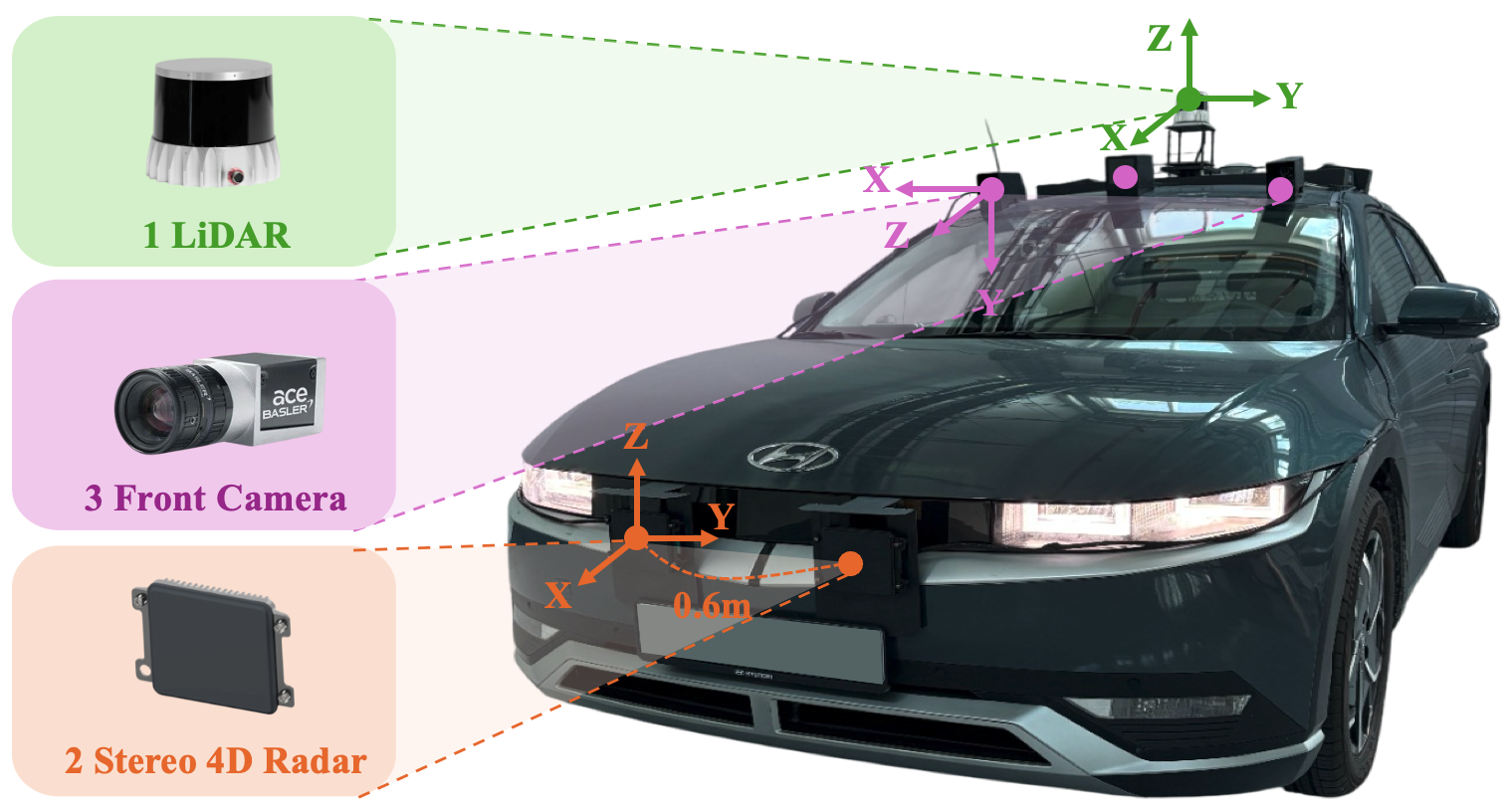}
    \caption{Sensor configuration for the stereo 4D Radar dataset.}
    \label{fig:6}
\end{figure}

To collect stereo 4D Radar data, we equipped a test vehicle with three types of sensors, as illustrated in Fig.~\ref{fig:6}. Two 4D Radars were mounted on the front bumper with a fixed baseline of 0.6\,m. Each Radar operates in the 76-79\,GHz band and is equipped 
with 12 transmit (Tx) and 16 receive (Rx) antennas.
In addition, a 64-channel long-range LiDAR was installed at the center of the vehicle roof, and three forward-facing cameras were mounted on the front roof section. The LiDAR and cameras were primarily used for data visualization and for generating annotations.

\subsection{Data Collection and Distribution}
The dataset was collected in Daejeon, South Korea, using the sensor platform described earlier. 
All data were recorded under clear daytime conditions and span a diverse set of driving environments, including 10 urban, 11 highway, and 16 suburban sequences. 
This composition provides representative autonomous-driving scenarios while maintaining consistent environmental conditions. For training and evaluation, 
the dataset is split into 18K frames for training and 4K frames for testing.

A total of 66K 3D bounding boxes were annotated for the \emph{Sedan} class. Specifically, an automated labeling pipeline\cite{autolabel} was first applied to the collected LiDAR data to obtain initial annotations, followed by manual refinement to improve label accuracy. The annotation focuses on a single class for two reasons. 
First, sedans constitute the majority of traffic participants in the collected sequences, providing sufficient data density and variability for reliable training and evaluation. 
Second, restricting the annotations to a dominant vehicle class avoids confounding factors arising from heterogeneous object types with differing Radar reflectivity patterns, enabling a clearer analysis of stereo 4D Radar-based velocity compensation and 3D detection performance. 
This design choice follows the common practice in prior Radar perception studies, where a primary vehicle class is analyzed first before extending to multi-class settings \cite{k-radar, rpfa, smurf}.

\bibliographystyle{IEEEtran}
\bibliography{ref}

\end{document}